\documentclass[letterpaper]{article} 
\usepackage{aaai2026}  
\usepackage{times}  
\usepackage{helvet}  
\usepackage{courier}  
\usepackage[hyphens]{url}  
\usepackage{graphicx} 
\usepackage{natbib}  
\usepackage{caption} 
\usepackage{algorithm}
\usepackage{algorithmic}

\usepackage{newfloat}
\usepackage{listings}
\DeclareCaptionStyle{ruled}{labelfont=normalfont,labelsep=colon,strut=off} 
\floatstyle{ruled}
\newfloat{listing}{tb}{lst}{}
\floatname{listing}{Listing}
\usepackage{url}
\usepackage{amsmath}
\usepackage{multirow}
\usepackage{verbatim}
\usepackage{enumitem}
\usepackage{amsfonts}
\usepackage{booktabs}

\title{SSTODE: Ocean-Atmosphere Physics-Informed Neural ODEs \\ for Sea Surface Temperature Prediction}
\author{
    Zheng Jiang\textsuperscript{\rm 1},
    Wei Wang\textsuperscript{\rm 1}\thanks{Corresponding author.},
    Gaowei Zhang\textsuperscript{\rm 1},
    Yi Wang\textsuperscript{\rm 2}
}
\affiliations{
    \textsuperscript{\rm 1}School of Artificial Intelligence, Beijing University of Posts and Telecommunications, Beijing, China\\
    \textsuperscript{\rm 2}School of Computer Science, Beijing University of Posts and Telecommunications, Beijing, China\\
    \{jiangzheng, weiwang, zhanggaowei\}@bupt.edu.cn, wang@cocolabs.org \\
}

\usepackage{bibentry}

\begin{document}
\maketitle
\begin{abstract}
Sea Surface Temperature (SST) is crucial for understanding upper-ocean thermal dynamics and ocean-atmosphere interactions, which have profound economic and social impacts. While data-driven models show promise in SST prediction, their black-box nature often limits interpretability and overlooks key physical processes. Recently, physics-informed neural networks have been gaining momentum but struggle with complex ocean-atmosphere dynamics due to 1) inadequate characterization of seawater movement (e.g., coastal upwelling) and 2) insufficient integration of external SST drivers (e.g., turbulent heat fluxes). To address these challenges, we propose SSTODE, a physics-informed Neural Ordinary Differential Equations (Neural ODEs) framework for SST prediction. First, we derive ODEs from fluid transport principles, incorporating both advection and diffusion to model ocean spatiotemporal dynamics. Through variational optimization, we recover a latent velocity field that explicitly governs the temporal dynamics of SST. Building upon ODE, we introduce an Energy Exchanges Integrator (EEI)—inspired by ocean heat budget equations—to account for external forcing factors. Thus, the variations in the components of these factors provide deeper insights into SST dynamics. Extensive experiments demonstrate that SSTODE achieves state-of-the-art performances in global and regional SST forecasting benchmarks. Furthermore, SSTODE visually reveals the impact of advection dynamics, thermal diffusion patterns, and diurnal heating-cooling cycles on SST evolution. These findings demonstrate the model’s interpretability and physical consistency.
\end{abstract}

\begin{links}
    \link{Code}{https://github.com/nicezheng/SSTODE-code}
\end{links}

\section{Introduction}
Covering 75\% of Earth's surface, the ocean regulates climate through ocean-atmosphere interactions governing energy exchange and hydrological cycles, largely governed by Sea Surface Temperature (SST) variability \citep{huang2005}. SST field constrains upper-ocean circulation and thermal structure across daily to decadal scales, modulating air-sea energy exchange \citep{garcia2021overview}. As the dominant climate fluctuation of this coupling, ENSO affects the global climate and disrupts normal weather patterns \citep{ham2019deep}. Accurate SST predictions at various time-scale are thus crucial for scientific and socioeconomic applications. 

\begin{figure}[t]
    \centering
    \includegraphics[width=\linewidth]{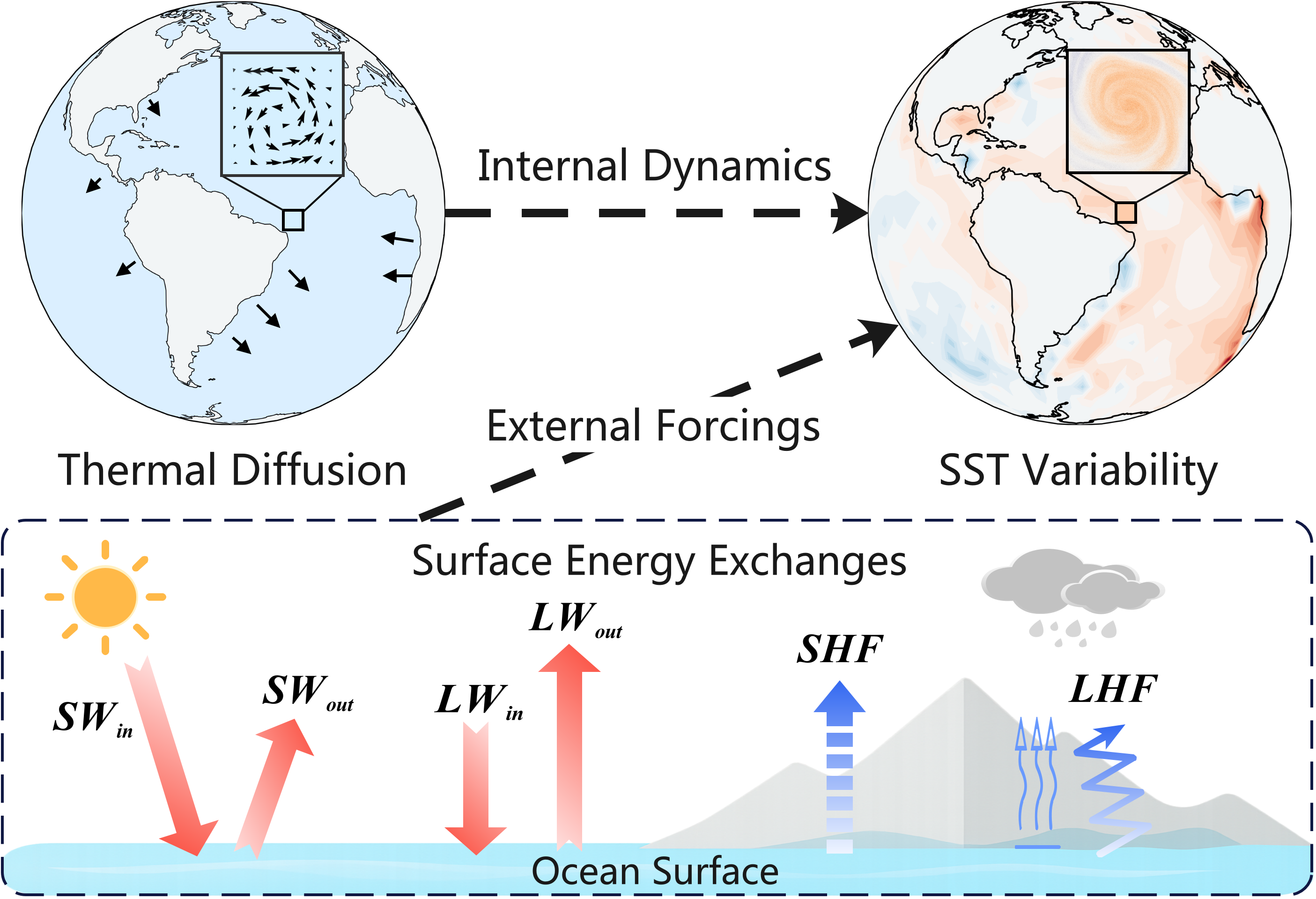}
    \caption{
    Two primary mechanisms govern SST variability (red: warming, blue: cooling): internal ocean dynamics (coupled advection-diffusion dynamics) and external energy exchanges. Among them, thermal diffusion is particularly crucial since it models heat spread driven by seawater movements like subgrid-scale eddies (see zoomed-in box), and coastal upwelling due to boundary effects (see black arrows). External energy exchanges mainly include shortwave radiation (SW), longwave radiation (LW), sensible heat flux (SHF), and latent heat flux (LHF).
    }
    \label{fig:intro}
\end{figure}
SST prediction has traditionally relied on coupled ocean-atmosphere models that numerically solve fluid dynamics equations, e.g., Navier-Stokes~\citep{singh2022review, ma2021multi}. While offering physical fidelity, they suffer from high computational cost. Deep learning models--from CNNs to Transformers--have recently advanced spatiotemporal SST forecasting by modeling complex ocean systems~\citep{tan2023openstl, shi2024oceanvp}. For example, Wenhai~\citep{cui2025forecasting} is a ViT-based model pretrained on 25 years of reanalysis data, enabling skillful multi-day SST forecasting (1-10 day lead times). Though predictive accuracy is essential, the omission of explicit fine-scale thermodynamic processes in these models compromises both physical consistency and interpretability--a fundamental requirement in ocean science. Furthermore, operating at fixed intervals (e.g., daily or weekly) limits their ability to capture multiscale ocean-atmosphere interactions that evolve across varying timescales.

Recently, Physics-Informed Neural Networks (PINNs) have emerged to incorporate physical laws into model training, preventing physically implausible predictions~\citep{li2024physics, li2024deepphysinet}. A typical strategy involves encoding partial
differential equations into Neural ODE frameworks~\citep{verma2024climode, xiang2024agc}.
However, the existing physical-informed architectures may not be suitable for SST prediction due to the following inherent challenges. 
(1) \textbf{Inadequate Characterization of Seawater Movement.} While the advection equations are typically the basis of existing Neural ODE models for atmospheric transport, SST evolution is governed by coupled advection-diffusion dynamics. As depicted in Fig.~\ref{fig:intro}, this diffusion component is essential to model subgrid-scale heat transfer processes associated with seawater movements, such as turbulence mixing, mesoscale eddies, or coastal upwelling. In particular, due to boundary effects, coastal upwelling generates sharp temperature gradients near coastlines and leads to numerical oscillation during model training \cite{salois2022coastal}. 
(2) \textbf{Insufficient Integration of External SST Drivers.} In the upper ocean layers, due to surface energy exchanges, the external factors (see Fig.~\ref{fig:intro}) further drive changes in SST tendency. Ignoring such forcings leads to biased predictions, particularly for diurnal heating--cooling cycles, regional warming trends, and seasonal anomalies.

To address the above challenges, we propose SSTODE, an ocean-atmosphere physics-informed Neural ODE framework for SST prediction. SSTODE formulates a continuous-time Neural ODE grounded in the advection--diffusion equation governing oceanic heat transport. Specifically, we incorporate a diffusion term into the Neural ODEs by applying the Laplacian of the SST field \citep{o2012operational}, scaled by a learnable thermal diffusivity. The Laplacian operator explicitly captures the spatial curvature of SST fields induced by subgrid-scale ocean dynamics, with particular effectiveness in resolving sharp thermal gradients around coastal boundary regions. The learned diffusivity dynamically modulates diffusion strength, smoothing abrupt SST variations in regions with strong thermal gradients. The combination effectively models seawater movement and enhances the numerical stability of ODE, ultimately generating physically consistent and accurate SST predictions. Building upon the Neural ODEs, we introduce an Energy Exchanges Integrator (EEI, a time-dependent network) to further account for external SST drivers in the upper open ocean. Derived from the ocean heat budget equation~\cite{vijith2020closing}, EEI incorporates four physically grounded surface flux factors (i.e., shortwave radiation, longwave radiation, latent heat flux, and sensible heat flux) from ERA5 reanalysis data to quantify their collective impact on SST dynamics over the forecast horizon. Comprehensive experiments on SST forecasting benchmark and visualizations substantiate the effectiveness and interpretability of SSTODE.

Our main contributions are as follows:
\begin{itemize}[noitemsep,left=0pt, itemsep=0pt, topsep=0pt]
    \item We propose SSTODE, a continuous-time Neural ODE framework that explicitly models coupled advection-diffusion processes for SST spatiotemporal prediction. 
    
    \item The EEI is developed that integrates surface heat fluxes, improving ODE solutions by accounting for air-sea energy exchanges--enhancing open-ocean predictions.
    
    \item Our model achieves state-of-the-art performance on global and regional SST forecasting benchmarks, revealing three physical mechanisms behind SST evolution: 1) coherent advection flows; 2) boundary-aware diffusion patterns; and 3) external forcing of diurnal variations. 
\end{itemize}

\section{Related Work}
\subsection{Ocean Parameters Prediction Models}
Numerical ocean models traditionally simulate ocean dynamics using physical equations but suffer from high computational cost and limited adaptability to dynamic external forcings \citep{veeresha2021numerical}.
Deep models for spatiotemporal forecasting, initially developed for computer vision tasks such as video and traffic prediction, have been extended to SST prediction using both recurrent-based (e.g., ConvLSTM~\citep{lin2020self}, Transformers~\citep{tang2024vmrnn}) and recurrent-free architectures (e.g., SimVPv2~\citep{tan2025simvpv2}, COTERE~\citep{shi2024oceanvp}). Large models have recently emerged as a promising paradigm in ocean and climate science, e.g., AI-GOMS \citep{xiong2023ai}, XiHe \citep{wang2024xihe}, and WenHai \citep{cui2025forecasting}, which leverage massive datasets and millions of parameters to improve prediction accuracy. While these DL models achieve impressive performance, their lack of explicit physical constraints limits their interpretability and thermodynamic consistency. 
Meanwhile, they are typically trained on reanalysis data sampled at fixed, discrete time intervals (e.g., daily), overlooking the fact that the seawater continuously transports heat and redistributes mass while surface fluxes add or remove energy. This discreteness violates mass conservation and introduces approximation errors. In contrast, our work focuses on ensuring physical interpretability by preserving intermediate states and enabling earlier anomaly detection through continuous-time forecasting.

\begin{figure*}[!t]
  \centering
  \includegraphics[width=\textwidth]{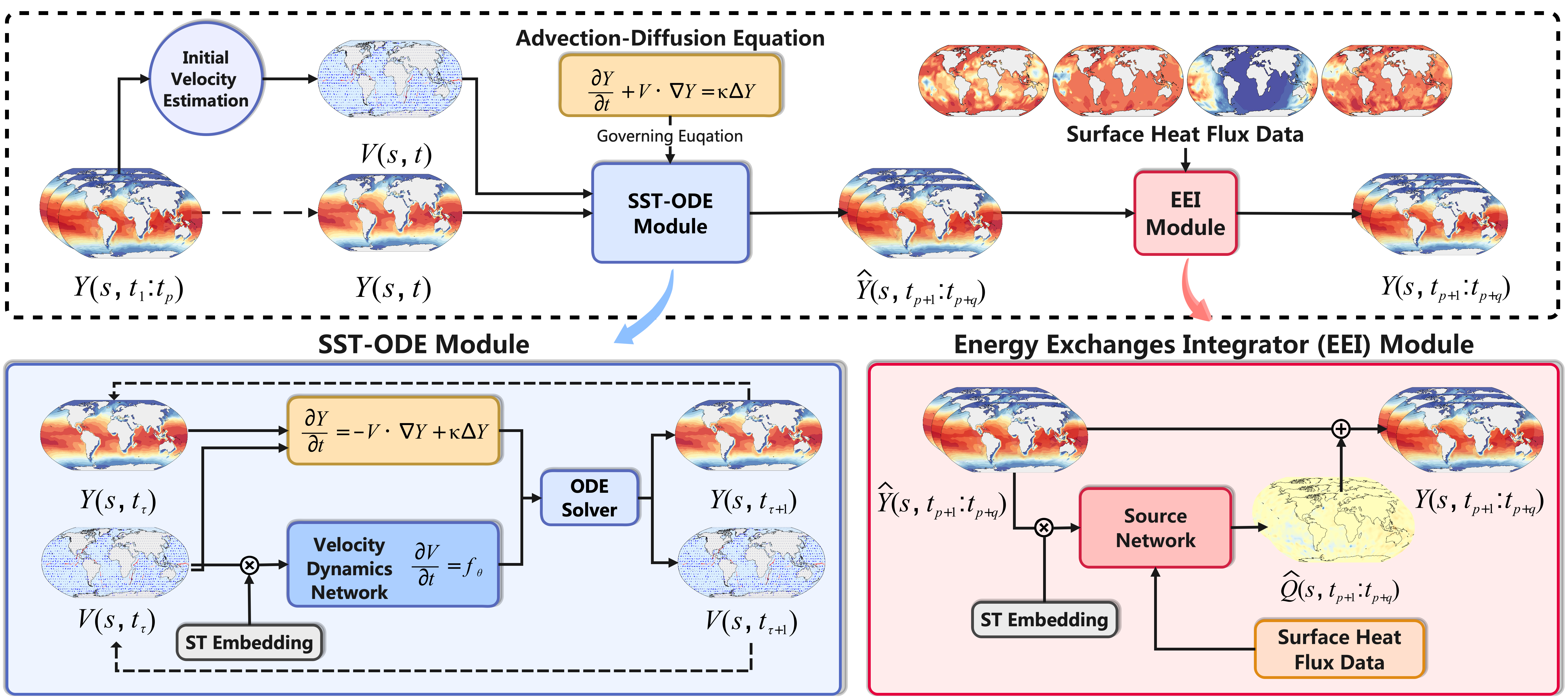}
  \caption{
  SSTODE framework. It comprises three modules: (1) Initial Velocity Estimation infers latent initial velocity from past SST by solving a PDE-constrained inverse problem; (2) SST-ODE integrates the advection–diffusion equation using Neural ODEs to prediction SST and latent velocity over continuous time; and (3) Energy Exchange Integrator (EEI) refines predictions using surface heat flux data via a learned source network. Spatiotemporal Embeddings (ST Embedding) encode position and time context. $\otimes$: concatenation. $\oplus$: element-wise addition.}
  \label{fig:framework}
\end{figure*}

\subsection{Physic-Informed Neural Networks}
Recent PINNs aim to combine data-driven learning with physical priors to improve prediction accuracy and physical plausibility \citep{xu2024generalizing, li2024deepphysinet}. A straightforward way is to embed the governing partial differential equations (PDEs) in the loss function to enforce physical laws~\citep{ghosh2023rans}. However, these methods often require full observation of key variables to close the equations, which is challenging for advection-diffusion systems because the requisite velocity fields are typically unavailable from satellite observations \citep{sun2020physics}.

Neural ODEs extend PINNs by parameterizing time derivatives with neural networks and integrating dynamics from initial conditions in continuous time~\citep{chen2018neural, chu2024adaptive}. This framework has shown success in modeling physical systems such as fluid dynamics~\citep{rojas2021reduced, tian2025air} and climate processes~\citep{hwang2021climate, verma2024climode}. For example, ClimODE embeds the advection equation to model atmospheric spatiotemporal patterns, achieving promising results in weather and climate prediction \citep{verma2024climode}. However, these methods often assume idealized advection in unbounded, compressible atmospheric settings, making them suboptimal for ocean modeling. Specifically, they can not adequately capture key ocean-specific dynamics: (1) intense thermal diffusion near coastlines from turbulence and eddies, (2) incompressible fluid assumptions unique to seawater, and (3) external forcings of surface energy exchange (e.g., radiative and turbulent heat fluxes). Our model addresses this limitation by explicitly integrating coupled advection-diffusion and surface energy exchange patterns into Neural ODE, enabling improved accuracy and realism of SST modeling in the upper open ocean.

\section{Method}

\subsection{Overview}
\subsubsection{Problem Definition}
We aim to predict the spatiotemporal evolution of SST distribution over the global ocean, represented by a 2D spatial domain $\Omega = [-90, 90] \times [-180, 180]$, with latitude and longitude defining the horizontal coordinates. Let $Y(s, t) \in \mathbb{R}$ denote the SST at location $s = (x, y) \in \Omega$ and time $t$. Given a sequence of $p$ past SST fields ${Y(s, t_1), \dots, Y(s, t_p)}$ sampled at uniform intervals $\Delta t$, the goal is to predict $q$ future fields ${Y(s, t_{p+1}), \dots, Y(s, t_{p+q})}$. We frame this task as a continuous-time neural dynamical system, initialized with the last observed field $Y(s, t_p)$ and integrated forward to generate future SST fields.

\subsubsection{Overall framework} 
As shown in Fig.~\ref{fig:framework}, SSTODE comprises three main components: the Initial Velocity Estimation, the SST-ODE Module, and the Energy Exchange Integrator (EEI) Module. Given past SST observations $Y(s, t_1{:}t_p)$, the Initial Velocity Estimation module infers the initial velocity field $\mathbf{V}(t_p)$ to initialize the ODE system. The SST-ODE Module then solves a continuous advection-diffusion ODE, integrating from $Y(s, t_p)$ with velocity $\mathbf{V}(t_p)$ to produce predictions $\hat{Y}(s, t_{p+1{:}p+q})$ iteratively. This module captures both large-scale advection and subgrid-scale thermal diffusion in a unified, physics-driven manner. To account for external forcings, the EEI Module integrates surface heat flux data through a mixed-layer source network and adds their effects to the ODE outputs, yielding the final SST forecast $Y(s, t_{p+1{:}p+q})$.

\subsection{Physical Dynamics Modeling}
\subsubsection{Physical Equations}
We formulate SST evolution using the advection–diffusion equation, which captures the core physical processes underlying spatiotemporal SST dynamics. This PDE models temporal SST changes driven by horizontal transport (advection) and subgrid-scale dynamics (diffusion), and is defined point-wise over the spatial domain as follows (Eq.~\ref{eq:pde}):
\begin{equation}   
    \begin{aligned}
    \underbrace{ \frac{\partial Y(s, t)}{\partial t} }_{\substack{\text{Temporal} \\ \text{Derivatives}}} 
    + \underbrace{ \mathbf{V}(s, t) \cdot \nabla Y(s, t) }_{\text{Advection}} 
    &= 
    \underbrace{ \kappa \Delta Y(s, t) }_{\text{Diffusion}} .
    \end{aligned}
    \label{eq:pde}
\end{equation}
\begin{itemize}[noitemsep,left=0pt, itemsep=0pt, topsep=0pt]
    \item \textbf{Advection term.} $\mathbf{V}(s, t) \cdot \nabla Y(s, t)$ represents the horizontal heat transport by ocean surface currents. Here, $\mathbf{V}(s, t) \in \mathbb{R}^2$ denotes the 2D velocity field consisting of zonal ($V_x$) and meridional ($V_y$) components, and $\nabla$ is the spatial gradient operator.
    \item \textbf{Diffusion term.} $\kappa \Delta Y(s, t)$ is subgrid-scale thermal diffusion caused by unresolved processes such as turbulence and mesoscale eddies. The scalar coefficient $\kappa \in \mathbb{R}^+$ controls the strength of diffusion, and $\Delta$ denotes the spatial Laplacian operator.
\end{itemize}

\subsubsection{ODE Dynamics}  
To enable continuous-time modeling with neural networks, we reformulate the advection–diffusion PDEs as a system of first-order ODEs using the Method of Lines (MOL)~\cite{iakovlev2021learning}. 
For notational simplicity, we omit the spatial coordinate $s$ in $Y(t)$ and $\mathbf{V}(t)$; all spatial operators $\nabla$ and $\Delta$ are applied over the discretized grid. The resulting continuous-time ODE integration is given in Eq.~\ref{eq:ode}:
\begin{equation}
\begin{aligned}
Y(t) &= Y(t_0) + \int_{t_0}^{t} \left[ - \mathbf{V}(\tau) \cdot \nabla Y(\tau) + \kappa \Delta Y(\tau) \right] d\tau, \\
\mathbf{V}(t) &= \mathbf{V}(t_0) + \int_{t_0}^{t} \dot{\mathbf{V}}(\tau)\, d\tau,
\end{aligned}
\label{eq:ode}
\end{equation}
where $\tau \in \mathbb{R}$ denotes continuous integration time, allowing predictions at arbitrary time resolutions without relying on fixed discrete steps. Such continuous integration is performed across multiple time intervals during the system’s evolution. Integration begins at $t_0$, the last observed timestep. The first equation describes SST evolution via the advection-diffusion equation, while the second defines the velocity field’s evolution through temporal derivative $\dot{\mathbf{V}}(\tau)$. These formulations underpin the framework presented in the subsequent sections.

\subsection{Initial Velocity Estimation}
The Neural ODE requires an initial condition to initiate integration, but the initial velocity field $\mathbf{V}(t_0)$ in Eq.~\ref{eq:ode} is typically unavailable in satellite data~\citep{sun2020physics}. We adopt the preprocessing approach from~\citet{verma2024climode} and estimate $\mathbf{V}(t_0)$ by solving a PDE-constrained inverse problem. Specifically, we first approximate the temporal derivative $\frac{\partial Y(t_0)}{\partial t}$ from past $p$ SST observations using cubic spline interpolation, yielding a smooth and differentiable proxy ground truth of local SST variation. We then optimize a learnable latent field $\hat{\mathbf{V}}_\theta(t_0) \in \mathbb{R}^{2 \times H \times W}$ by minimizing the loss of MSE between the interpolated and ODE-computed derivatives under the advection–diffusion equation, with a spatial smoothness regularization term.
\begin{equation}
\begin{aligned}
    \hat{\mathbf{V}}_{\theta}(t_0) = \arg\min_{\mathbf{V}_{\theta}(t)} 
\left\{ \| R(t_0) \|^2 + \alpha \| \mathbf{V}_{\theta}(t) \|^2 \right\},
\label{eq:vel_opt}
\end{aligned}
\end{equation}

\begin{equation}
\begin{aligned}
    R(t) = \frac{\partial Y(t)}{\partial t} + \mathbf{V}_{\theta}(t) \cdot \nabla Y(t) - \kappa_{\theta} \Delta Y(t),
\label{eq:residual_term}
\end{aligned}
\end{equation}
where $\kappa_{\theta}$ is a learnable scalar diffusion coefficient constrained via softplus, and $\alpha$ controls the strength of spatial smoothing. We also place an RBF-kernel Gaussian prior on $\mathbf{V}_\theta(t)$ to encourage coherence. See Appendix A.1 for details and an initial condition robustness test.

\subsection{SST-ODE Module}
The SST-ODE Module jointly models the continuous-time evolution of SST and latent velocity by Neural ODEs grounded in the advection–diffusion equation. As shown in Fig.~\ref{fig:framework}, the module solves the advection–diffusion equation and explicitly captures both large-scale advection and subgrid-scale thermal diffusion. Initialized with estimated velocity field $\mathbf{V}(t_0)$ and SST field $Y(t_0)$, the integration progresses in continuous time. The dynamics are split into two ODEs: one for SST evolution, and one for latent velocity evolution.

\subsubsection{Temporal SST Evolution}

Given the velocity field $\mathbf{V}(t)$, the SST temporal derivative is computed using the advection–diffusion equation. At each continuous integration time $\tau$, this yields:
\begin{equation}
    \frac{\partial Y(\tau)}{\partial \tau} =  - \mathbf{V}(\tau) \cdot \nabla Y(\tau) + \kappa_{\theta} \Delta Y(\tau),
\label{eq:pde_tau}
\end{equation}
where $\kappa_{\theta}$ is parameterized similarly to the initial velocity estimation. The resulting derivative is integrated from the initial SST state using a Neural ODE solver, yielding a continuous and differentiable prediction trajectory. This formulation supports flexible forecasting over arbitrary horizons while preserving a physically interpretable decomposition of SST dynamics into advection and diffusion processes.

\subsubsection{Temporal Velocity Evolution} 
Unlike SST, directly modeling velocity evolution is challenging due to the lack of pressure observations and the nonlinearity inherent of the Navier-Stokes equations \citep{hoang2010navier}. To address this, we parameterize the complex velocity evolution $\dot{\mathbf{V}}(\tau)$ as a neural function learned during forward ODE integration. Inspired by the influence of temperature gradients on fluid motion through buoyancy and density-driven forces~\citep{willeit2024generalized}, we condition the velocity dynamics on both the SST field and its spatial gradients. At each integration time $\tau$, the temporal derivative of velocity is defined as:
\begin{equation}
\begin{aligned}
\frac{\partial \mathbf{V}(\tau)}{\partial \tau} = \dot{\mathbf{V}}(\tau) = f_v \left( \mathbf{V}(\tau), Y(\tau), \nabla Y(\tau), \phi(s, \tau) \right),
\end{aligned}
\end{equation}
where $f_v$ is a neural velocity dynamics network, and $\phi(s, \tau)$ denotes the spatiotemporal embedding (See Appendix A.2). To capture both local and global dependencies, $f_v$ adopts a hybrid architecture network combining ResNet blocks and attention modules. Similar to SST dynamics, the velocity field $\mathbf{V}(t)$ is continuously updated over time by integrating $\dot{\mathbf{V}}(t)$ using a Neural ODE solver.

\subsection{Energy Exchange Integrator Module}
To incorporate external drivers of SST dynamics, the EEI Module estimate surface energy exchanges (the \textbf{Source Term} $Q(s, t)$), and refines the ODE-based outputs accordingly. We begin by outlining the external data sources, followed by our estimation and integration strategy.

\subsubsection{External Forcing Variables}
Inspired by the ocean mixed-layer heat budget~\citep{shu2024improved}, which relates surface energy exchanges to net energy fluxes $Q_{\text{net}}$ at the air–sea interface, we adopt the following formulation:
\begin{equation}
    Q_{\text{net}} = \frac{Q_{\text{LW}} + Q_{\text{SW}} + Q_{\text{LHF}} + Q_{\text{SHF}}}{\rho c_p h},
    \label{eq:heatbudge}
\end{equation}
where $\rho$, $h$, and $c_p$ are the mean density, the mixed layer depth, and the specific heat capacity of sea water, respectively. The numerator includes four dominant fluxes: shortwave ($Q_{\text{SW}}$) and longwave radiation ($Q_{\text{LW}}$), latent ($Q_{\text{LHF}}$), and sensible heat flux ($Q_{\text{SHF}}$). These four components constitute the dominant forms of energy exchange at the ocean surface and are considered essential to SST evolution. 

\subsubsection{Estimation and Integration}
We construct the external inputs $H(s, t) = \{Q_{LW}, Q_{SW}, Q_{LHF}, Q_{SHF}\}$ by concatenating four surface heat flux variables. Since future $H(s, t)$ is unavailable during inference, we extend $H(s, t_0)$ across the prediction horizon and concatenate it with the predicted SST fields $\hat{Y}(s, t_{p+1{:}p+q})$ and spatiotemporal embeddings $\phi$ (see Appendix A.2). These inputs are fed into a time-dependent neural network $f_s$ (Source Network in Fig.~\ref{fig:framework}) to estimate the sequence of future source terms:
\begin{equation}
\hat{Q}(s, t_{p+1{:}p+q}) = f_s \left( H(s, t_0), \hat{Y}(s, t_{p+1{:}p+q}), \phi \right).
\end{equation}
This temporally-aware design captures both the impact of external forcing on SST and the feedback effects from SST to surface energy exchange, implicitly modeling short-term feedback effects driven by thermodynamic coupling at the air–-sea interface. The estimated source term $\hat{Q}(s, t)$ is then added as a correction to the ODE-based forecast:
\begin{equation}
Y(s, t_{p+1{:}p+q}) = \hat{Y}(s, t_{p+1{:}p+q}) + \hat{Q}(s, t_{p+1{:}p+q}).
\end{equation}
This formulation explicitly integrates external energy inputs into the forecasting pipeline, allowing the model to adjust its forecast based on real-world environmental variability.

\section{Experiment}
\subsection{Dataset}
We evaluate our method on the OceanVP~\citep{shi2024oceanvp} and ERA5~\citep{hersbach2020era5} benchmark. OceanVP benchmark is built on HYCOM reanalysis~\citep{chassignet2007hycom}, providing 3-hourly global surface temperature, salinity, and velocity fields from 1994 to 2015 at $5.625^\circ$ resolution. We follow the standard split: 1994–2013 (train), 2014 (val), and 2015 (test). ERA5 offers hourly global reanalysis data from 1940 onward at $0.25^\circ$ resolution. We extract SST as the target and use four heat flux components—shortwave, longwave, latent, and sensible fluxes—to drive the EEI module. To avoid potential long-term ocean memory effects \citep{ham2019deep}, we introduce a 4-year gap by training on 2000 to 2015, validating on 2016, and testing on 2021 to 2022.

\begin{table*}[t]
\renewcommand{\arraystretch}{0.8}
\small
\centering
\begin{tabular}{l|cc|cc|cc|cc|cc|cc}
\toprule
\multirow{2}{*}{Model}
& \multicolumn{6}{c|}{OceanVP}
& \multicolumn{6}{c}{ERA5} \\
& \multicolumn{2}{c}{q=5} & \multicolumn{2}{c}{q=7} & \multicolumn{2}{c|}{q=12}
& \multicolumn{2}{c}{q=5} & \multicolumn{2}{c}{q=7} & \multicolumn{2}{c}{q=12} \\
& MSE ↓ & ACC ↑ & MSE ↓ & ACC ↑ & MSE ↓ & ACC ↑
& MSE ↓ & ACC ↑ & MSE ↓ & ACC ↑ & MSE ↓ & ACC ↑ \\
\midrule
ConvLSTM   & 0.0781 & 0.9985 & 0.1091 & 0.9980 & 0.1758 & 0.9967
           & 0.1745 & 1.0000 & 0.1879 & 1.0000 & 0.2295 & 1.0000 \\
MAU        & 0.0627 & 0.9988 & 0.0792 & 0.9985 & 0.1143 & 0.9978
           & 0.0242 & 1.0000 & 0.0326 & 1.0000 & 0.0508 & 1.0000 \\
PredRNNv2  & 0.1081 & 0.9980 & 0.1251 & 0.9977 & 0.1552 & 0.9972
           & 0.1892 & 1.0000 & 0.2081 & 1.0000 & 0.2007 & 1.0000 \\
\midrule
TAU        & 0.0576 & 0.9989 & 0.0756 & 0.9985 & 0.1092 & 0.9979
           & 0.0242 & 1.0000 & 0.0326 & 1.0000 & 0.0508 & 1.0000 \\
SimVPv2    & 0.0612 & 0.9988 & 0.0815 & 0.9984 & 0.1221 & 0.9977
           & 0.0227 & 1.0000 & 0.0295 & 1.0000 & 0.0450 & 1.0000 \\
COTERE     & 0.0624 & 0.9988 & 0.0778 & 0.9984 & 0.1062 & 0.9979
           & 0.0284 & 1.0000 & 0.0335 & 1.0000 & 0.0612 & 1.0000 \\
\midrule
ClimODE    & 0.0622 & 0.9988 & 0.0705 & 0.9986 & 0.1004 & 0.9981
           & 0.0196 & 1.0000 & 0.0258 & 1.0000 & 0.0408 & 1.0000 \\
\textbf{SSTODE} 
           & \textbf{0.0527} & \textbf{0.9990} & \textbf{0.0638} & \textbf{0.9987} & \textbf{0.0954} & \textbf{0.9981}
           & \textbf{0.0180} & 1.0000 & \textbf{0.0232} & 1.0000 & \textbf{0.0349} & 1.0000 \\
\bottomrule
\end{tabular}
\caption{Global forecasting results on OceanVP and ERA5 at prediction steps $q$ = 5, 7, and 12. The evaluation metrics are MSE ↓ and ACC ↑, where ↓ indicates lower is better, and ↑ indicates higher is better. Bold indicates the best result.}
\label{tab:global-results}
\end{table*}

\subsection{Baselines}
We benchmark our method against representative spatiotemporal forecasting baselines from three categories: \textbf{Recurrent-based}: ConvLSTM \citep{shi2015convolutional}, MAU \citep{chang2021mau}, and PredRNNv2 \citep{wang2022predrnn} utilize recurrent structures to capture temporal dependencies. \textbf{Recurrent-free}: TAU \citep{tan2023temporal}, SimVPv2 \citep{tan2025simvpv2}, and CORETE \citep{shi2024oceanvp} employ parallel architectures for improving efficiency and scalability. \textbf{Physics-Informed}:
ClimODE~\citep{verma2024climode} incorporates the advection equation within a Neural ODE framework without requiring velocity inputs. Neural operator (NO) models learn field-to-field mappings without enforcing explicit physical constraints and thus compared separately in Appendix A.9.

\subsection{Global Ocean Forecasting Results}
We evaluate SSTODE for global SST forecasting on OceanVP and ERA5, comparing against representative spatiotemporal baselines. To balance computational cost and support long-term forecasting, all data are downsampled to $5.625^\circ$ resolution, 6-hour intervals, and standardized. Models observe 3 past snapshots and predict $q$ = 5, 7, 12 future steps (30h, 42h, 72h). We report MSE, and Anomaly Correlation Coefficient (ACC) after de-normalization to assess both short-term variability and long-range dynamics. Table~\ref{tab:global-results} shows that SSTODE consistently outperforms all baselines on both OceanVP and ERA5 across all horizons ($q$ = 5, 7, 12). Compared to both recurrent-based and recurrent-free baselines, SSTODE leverages embedded physical priors for better performance under limited observations. It also outperforms the physics-informed ClimODE by jointly modeling thermal diffusion and surface energy exchange.

\subsection{Regional Ocean Forecasting Results}

\begin{table}[t]
\renewcommand{\arraystretch}{0.8}
\setlength{\tabcolsep}{3pt}
\small
\centering
\begin{tabular}{l|cc|cc|cc}
\toprule
\multirow{2}{*}{Model} 
& \multicolumn{2}{c|}{E.P.} 
& \multicolumn{2}{c|}{N.A.}
& \multicolumn{2}{c}{S.O.} \\
& q=5 & q=12
& q=5 & q=12
& q=5 & q=12 \\
\midrule
ConvLSTM   & 0.5578 & 1.1256 & 1.4052 & 2.9191 & 0.4766 & 1.1772 \\
MAU        & 0.5590 & 0.8755 & 1.1125 & 1.8058 & 0.3504 & 0.7103 \\
PredRNNv2  & 0.9216 & 1.2223 & 1.9601 & 2.6103 & 0.6236 & 0.9231 \\
\midrule
TAU        & 0.5736 & 0.9200 & 1.0621 & 1.8038 & 0.3014 & 0.6135 \\
SimVPv2    & 0.5160 & 0.9612 & 1.1709 & 1.9543 & 0.3051 & 0.6946 \\
COTERE     & 0.5736 & 0.8653 & 1.1118 & 1.7515 & 0.3259 & 0.6183 \\
\midrule
ClimODE    & 0.6064 & 0.8466 & 1.1234 & 1.7046 & 0.3098 & 0.5716 \\
\textbf{SSTODE} 
           & \textbf{0.5157} & \textbf{0.7473} & \textbf{0.9432} & \textbf{1.6445} & \textbf{0.2641} & \textbf{0.5597} \\
\bottomrule
\end{tabular}
\caption{Regional forecasting results (MSE ↓) on OceanVP across Equatorial Pacific (E.P.), Northwest Atlantic (N.A.), and Southern Ocean (S.O.) at forecast steps $q$ = 5, 12. Bold indicates the best result per column. MSE is scaled $\times 1000$ for better readability.}
\label{tab:regional-results}
\end{table}

We evaluate our model on regional forecasting over three oceanographically diverse domains constrained by fixed spatial bounding boxes \citep{lee2022future, jiang2021impacts, kang2023global}:  
(1) the Equatorial Pacific ($5^\circ\text{S}$–$5^\circ\text{N}$, $160^\circ\text{E}$–$100^\circ\text{W}$), which captures tropical SST variability linked to ENSO dynamics;
(2) the Northwest Atlantic ($30^\circ\text{N}$–$50^\circ\text{N}$, $80^\circ\text{W}$–$40^\circ\text{W}$), dominated by the Gulf Stream mesoscale activity; and
(3) the Southern Ocean ($50^\circ\text{S}$–$65^\circ\text{S}$, all longitudes), representing high-latitude regions sensitive to heat exchange and climate variability.
We use the same experimental setup as in the global evaluation. Table~\ref{tab:regional-results} shows that SSTODE consistently achieves notable gains across all regions and both horizons ($q$ = 5, 12). See Appendix A.5 for complete metrics.

\subsection{Visualization Analysis of Neural ODE Dynamics}
\begin{figure*}[t]
  \centering
  \includegraphics[width=\textwidth]{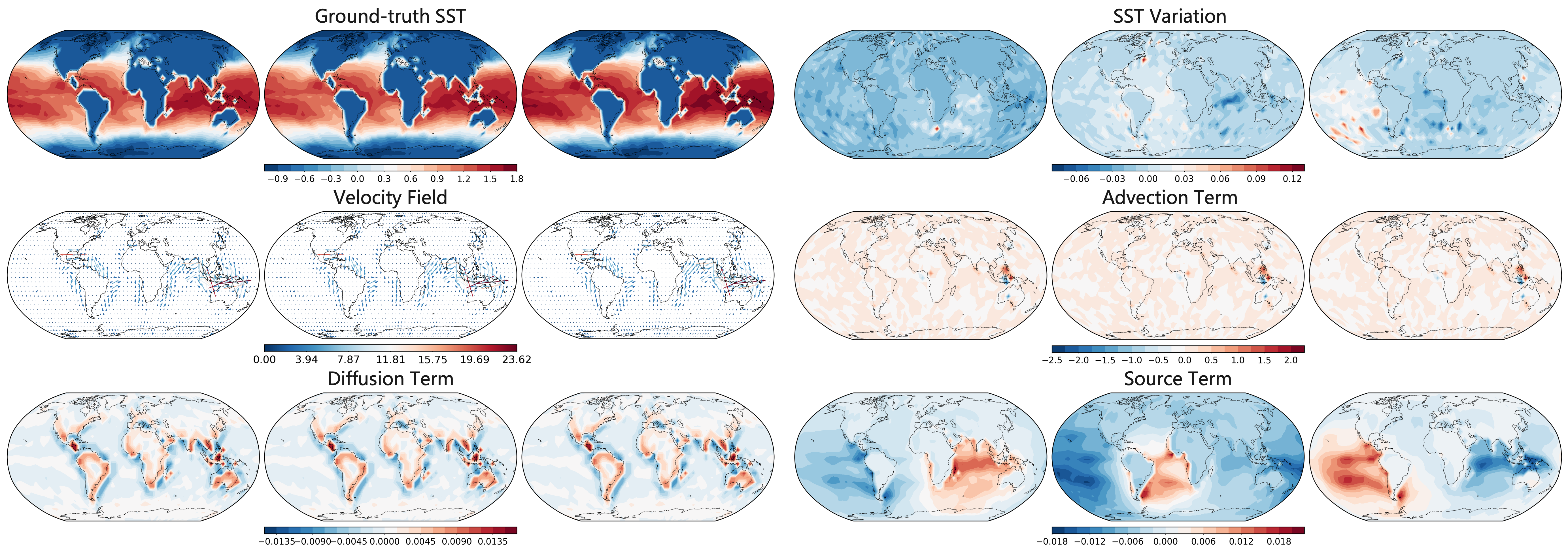}
  \caption{Visualization of intermediate dynamics across three time steps ($t$ = 1 to 3). Rows show: Ground-truth SST, SST Variation, Advection, Diffusion, Velocity Field, and Source Term. SSTODE decouples these components, yielding interpretable and physically grounded results across the forecast horizon.}
  \label{fig:ode-vis}
\end{figure*}

To further demonstrate the physical interpretability of SSTODE, we visualize intermediate components of the advection-diffusion over three forecast steps ($t=1$ to $t=3$) in Fig.~\ref{fig:ode-vis}. Specifically, we decompose the dynamics into:
(1) ground-truth SST $Y(t)$,
(2) observed SST variation $\Delta Y = Y(t+1) - Y(t)$.
(3) learned velocity field $\mathbf{V}(t)$,
(4) advection term $-\mathbf{V}(t) \cdot \nabla Y(t)$,
(5) diffusion term $\kappa \Delta Y(t)$, and
(6) source term $Q(t)$ estimated by external forcings. 
These visualizations illustrate how SSTODE decouples SST evolution into interpretable physical terms.
\begin{itemize}[noitemsep,left=0pt, itemsep=0pt, topsep=0pt]
    \item \textbf{Velocity Field \& Advection.} These fields reveal coherent large-scale circulations and localized high-magnitude structures. Red arrows highlight key advection zones (e.g., Western Pacific Warm Pool, Indonesian Throughflow, and Caribbean Current between North and South America), which align with known boundary current systems. This indicates that the model captures physically consistent velocity patterns essential for SST transport.
    \item \textbf{Diffusion term.} This component shows smooth and coherent spatial patterns across ocean regions, demonstrating that the diffusivity effectively smooths temperature gradients and models gradual heat transport while avoiding overfitting to local noise. Notably, the model effectively captures sharp gradients near coastlines, aligning with dynamic coastal processes such as coastal upwelling and boundary currents.
    \item \textbf{Source term.} The learned source term captures diurnal patterns aligned with radiative and turbulent fluxes, presenting strong correspondence with observed SST variations—particularly in regions undergoing day–night heating transitions. This consistency underscores the source term’s contribution to accurate forecasting and emphasizes the importance of modeling external energy exchanges for reliable SST prediction.
\end{itemize}

\subsection{Robustness Analysis}

\subsubsection{Ablation Study on Diffusion Term Design} 
We evaluate the effect of thermal diffusion in SST forecasting by removing the diffusion term and comparing three formulations of the diffusion coefficient $\kappa$ in Eq.~\ref{eq:ode}, ranging from fixed diffusivity (no parameterization) to spatially varying setups (high parameterization): (i) a fixed $\kappa$ to a constant ($\kappa=1$, as used in~\citep{vijith2020closing}), (ii) a learnable global scalar, and (iii) a learnable spatially-varying 2D map. Experiments on OceanVP with a 5-step forecast show that removing diffusion markedly degrades performance, while the global scalar yields the best results across all metrics (Table~\ref{tab:ablation-diffusion}). In practice, we observed that the 2D diffusivity map often fails due to over-parameterization. With only SST observed and velocity inferred latently, $\kappa$$\Delta T$ becomes unstable and partly interchangeable with advection term $-\mathbf{V}\cdot\nabla T$ in explaining SST tendencies. In contrast, the global scalar diffusivity acts as an effective regularizer stabilizing the ODE integration and capturing global diffusion.

\begin{figure}[!h]
  \centering
  \includegraphics[width=\linewidth]{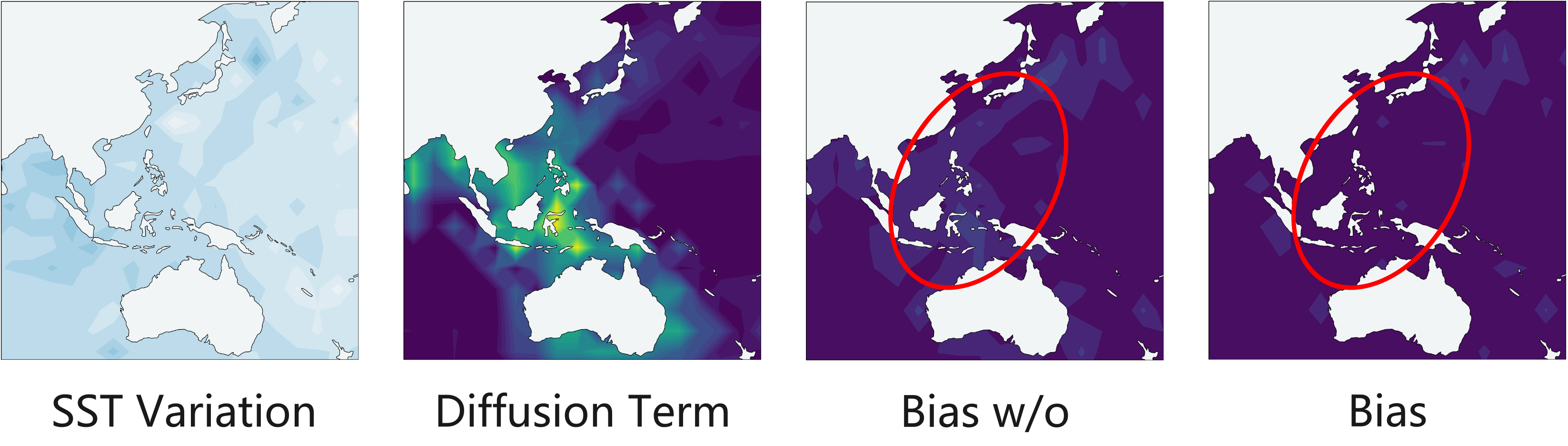}
  \caption{Impact of diffusion modeling on prediction bias. Left to right: SST variation, learned diffusion term with strong coastal responses, prediction bias without diffusion, and reduced bias with diffusion, especially in the red box.}
  \label{fig:diff-bias}
\end{figure}

Fig.~\ref{fig:diff-bias} reveals that regions with strong SST gradients, particularly near coastlines and straits in the Northwest Pacific, are highly sensitive to diffusion modeling. Excluding the diffusion term leads to numerical artifacts during and larger errors in these areas. In contrast, the diffusion term acts as a physical regularizer, enabling the model to capture sharp gradients and coastal dynamics (e.g., coastal upwelling) more accurately and stably during training.

\begin{table}[ht]
\renewcommand{\arraystretch}{0.8}
\centering
\begin{tabular}{lccc}
\toprule
Model Variant & MSE ↓ & MAE ↓ & ACC ↑ \\
\midrule
SSTODE w/o diffusion & 0.0596 & 0.1185 & 0.9989 \\
SSTODE Kappa (Fixed) & 0.0562 & 0.1158 & 0.9989 \\
SSTODE Kappa (2D) & 0.0578 & 0.1166 & 0.9989 \\
SSTODE & \textbf{0.0527} & \textbf{0.1107} & \textbf{0.9990} \\
\bottomrule
\end{tabular}
\caption{Diffusion term ablation. Comparison of diffusivity $\kappa$ parameterizations and removal of diffusion (w/o) cases at forecast step q=5.}
\label{tab:ablation-diffusion}
\end{table}

\subsubsection{Ablation Study on Source Term Design}  
\begin{table}[h]
\renewcommand{\arraystretch}{0.8}
\centering
\begin{tabular}{lccc}
\toprule
Model Variant & MSE ↓ & MAE ↓ & ACC ↑ \\
\midrule
SSTODE w/o Source & 0.0595 & 0.1198 & 0.9988 \\
SSTODE + SW & 0.0546 & 0.1109 & 0.9990  \\
SSTODE + LW & 0.0560 & 0.1135 & 0.9990  \\
SSTODE + LHF & 0.0559 & 0.1136 & 0.9990   \\
SSTODE + SHF & 0.0554 & 0.1173 & 0.9990  \\
SSTODE & \textbf{0.0527} & \textbf{0.1107} & \textbf{0.9990} \\
\bottomrule
\end{tabular}
\caption{Source term ablation. Performance at forecast step $q$ = 5 with incremental addition of surface heat flux component (SW, LW, LHF, SHF) in the EEI module.}
\label{tab:ablation-source}
\end{table}
We assess the contribution of each surface heat flux variable used in the EEI Module by incrementally adding SW, LW, LHF, and SHF. As shown in Table~\ref{tab:ablation-source}, adding the full source term leads to a substantial performance gain, underscoring the importance of energy forcing. Among individual terms, SW contributes the most, while all variables offer complementary benefits for accurate SST forecasting.

\subsubsection{Longer Horizon and Sparse Sampling}
We evaluate temporal generalization over a longer 7-day forecast horizon with both standard 6-hourly ($q$ = 28) and sparser 12-hourly ($q$ = 14) intervals on OceanVP. SSTODE demonstrates robust performance against strong baselines under both settings, as reported in Appendix A.6.

\subsubsection{Scalability to Higher Spatial Resolution}
We assess SSTODE's scalability to finer spatial grids by increasing the resolution to $2.8125^\circ$. Using SST and energy forcing variables extracted from ERA5, experiments in Appendix A.7 report that SSTODE maintains strong performance against representative baselines, demonstrating its potential to scale effectively to even finer resolutions.

\subsubsection{Model Parameters Comparison}
We compare model parameters under identical settings on a $2.8125^\circ$-resolution SST forecasting in Appendix A.8. SSTODE demonstrates superior parameter efficiency over both recurrent-based and recurrent-free baselines, benefiting from its autoregressive design. With a marginal increase in parameter count (4.62M vs 3.67M) over ClimODE, SSTODE achieves an $\sim$10\% improvement in forecasting accuracy--a favorable trade-off that remains highly acceptable for practical deployment.

\section{Conclusion}
SST prediction remains fundamentally important for understanding evolving ocean structure and its feedback with atmospheric circulation. This paper addresses the challenges of adapting Neural ODE framework for SST forecasting by incorporating explicit advection-diffusion constraints to model ocean dynamics and designing an EEI Module to account for surface heat fluxes. Extensive experiments demonstrate that our model achieves competitive performance, effectively capturing key physical processes such as advection, boundary-aware diffusion and diurnal variations. This work highlights the potential of combining physical priors with Neural ODEs for more interpretable and accurate ocean forecasting and also establishes it as a promising component for foundation models that support regional refinement and physically consistent downscaling. Future work will explore the extension to ocean subsurface thermohaline and multiscale dynamics.

\section{Acknowledgments}
This work is partially supported by National Natural Science Foundation of China under grants 62076232 and 62172049.

\bibliography{aaai2026}

\appendix
\newpage

\section{Appendix}
\setcounter{table}{4}
\setcounter{equation}{9}
\subsection{A.1 Initial Velocity Estimation}
\subsubsection{Details} We first approximate the temporal derivative $\frac{\partial Y(t_0)}{\partial t}$ using cubic spline interpolation over the past $p$ SST observations as proxy ground truth, yielding a smooth and differentiable estimate of local SST variation. We then parameterize the latent velocity field $\hat{\mathbf{V}}_\theta(t_0) \in \mathbb{R}^{2 \times H \times W}$ as a learnable tensor with zonal and meridional components, where $\theta$ denotes trainable neural network weights defined over the spatial grid. To estimate $\mathbf{V}(t_0)$, we jointly optimize $\hat{\mathbf{V}}_{\theta}(t_0)$ by minimizing the residual of the advection–diffusion equation along with an $\ell_2$ regularization. The final estimate $\mathbf{V}(t_0)$ is selected by minimizing the MSE between interpolated and those computed via ODE integration. For completeness, we restate equations here:
\begin{equation}
\begin{aligned}
    \hat{\mathbf{V}}_{\theta}(t_0) = \arg\min_{\mathbf{V}_{\theta}(t)} 
\left\{ \| R(t_0) \|^2 + \alpha \| \mathbf{V}_{\theta}(t) \|^2 \right\},
\label{eq:vel_opt}
\end{aligned}
\end{equation}
\begin{equation}
\begin{aligned}
    R(t) = \frac{\partial Y(t)}{\partial t} + \mathbf{V}_{\theta}(t) \cdot \nabla Y(t) - \kappa \Delta Y(t).
\label{eq:residual_term}
\end{aligned}
\end{equation}
The first term in Eq.~\ref{eq:vel_opt} penalizes deviations from the advection--diffusion equation by the residual $R(t)$ in Eq.~\ref{eq:residual_term}, while the second term enforces spatial coherence through $\ell_2$ regularization, weighted by a hyperparameter $\alpha$. To ensure physical plausibility and numerical stability, the diffusion coefficient $\kappa$ is modeled as a shared learnable scalar constrained to be positive through a softplus activation. Additionally, $\mathbf{V}_{\theta}(t)$ is regularized by a zero-mean Gaussian prior with a spatial covariance matrix defined by a radial basis function (RBF) kernel $K_{ij} = \text{rbf}(x_i, x_j; \alpha)$. This prior encourages spatial smoothness in $\mathbf{V}_{\theta}(t)$, with the kernel bandwidth $\alpha$ controlling its strength. In implementation, we set $\alpha = 1\text{e}{-7}$ and optimize the objective for 200 epochs using Adam with a learning rate of 2.

\subsubsection{Initial Condition Robustness Test} To assess the impact of initial velocity estimation stability on final results, we perform a series of initial velocity estimations by varying the number of optimization epochs from 50 to 400 (i.i., 50, 100, 200, 300, and 400). Each estimated velocity field is then used to perform downstream SST forecasting. Across these runs, we observe that convergence typically occurs within 200 epochs. We report results on the OceanVP dataset under a 6-hour interval forecasting setting with $q=5$ steps in Table~\ref{tab:v0-robustness}. The resulting MSE remains consistent across multiple initializations, with an average of $0.0542 \pm 0.0010$. This slight drop of performance and low variance demonstrates the reliability and reproducibility of the estimated initial condition despite differences in optimization trajectories.

\begin{table}[h!]
\centering
\small
\begin{tabular}{c|c}
\toprule
Epochs & MSE ↓ \\
\midrule
50   & 0.0556 \\
100  & 0.0548 \\
200  & 0.0527 \\
300  & 0.0539 \\
400  & 0.0538 \\
\midrule
\textbf{Mean ± Std} & \textbf{0.0542 ± 0.0010} \\
\bottomrule
\end{tabular}
\caption{Robustness of initial velocity estimation $V(t_0)$ across different optimization epochs on OceanVP (6-hour interval, $q$ = 5). Reported values are MSE, MAE and ACC. The last row shows the mean and standard deviation.}
\label{tab:v0-robustness}
\end{table}

\subsection{A.2 SpatialTemporal Embedding}
We integrate a spatiotemporal embedding (ST Embedding) into SSTODE to encode spatial coordinates and temporal patterns as auxiliary inputs, enhancing the modeling of spatiotemporal dynamics through positional priors. The spatial embedding $\phi_s$ encodes the latitude–longitude grid position $s = (x, y)$ of each spatial point using trigonometric and interaction functions:
\begin{equation}
\phi_s(s) = \left\{ \sin(k), \cos(k) \right\} \times \left\{ x, y, x \cdot y \right\}.
\end{equation}
The temporal embedding $\phi_t$ captures both daily and seasonal periodicity, leveraging trigonometric functions to model the cyclical nature of time:
\begin{equation}
\phi_t(t) = \left[ \sin(2\pi t), \cos(2\pi t), \sin\left( \frac{2\pi t}{365} \right), \cos\left( \frac{2\pi t}{365} \right) \right].
\end{equation}
We construct the final spatiotemporal embedding $\phi(s, t)$ by concatenating the spatial and temporal encodings, their element-wise interaction, and static geographic features:
\begin{equation}
\phi(s, t) = \left[ \phi_s(s), \phi_t(t), \phi_s(s) \odot \phi_t(t), lsm(s), oro(s) \right].
\end{equation}
Here, $\odot$ denotes the element-wise product between the spatial and temporal embeddings. The terms $lsm(s)$ and $oro(s)$ represent static land-sea mask and orography (elevation), respectively. The resulting embeddings are concatenated with raw physical variables and external data as additional inputs to the network.

\subsection{A.3 Implementation Details}
In the Initial Velocity Estimation module, we approximate the temporal derivative $\frac{\partial Y(t)}{\partial t}$ using \texttt{torchcubicspline} \footnote{\url{https://github.com/patrick-kidger/torchcubicspline}}, and compute spatial gradients via \texttt{torch.gradient}. The initial velocity $\mathbf{V}(t_0)$ is estimated by minimizing the PDE residual loss, with RBF-based spatial regularization ($\alpha = 1\text{e}{-7}$), optimized using Adam (lr=2, 200 epochs). We integrate the Neural ODEs using the \texttt{torchdiffeq} library \footnote{\url{https://github.com/rtqichen/torchdiffeq}} with the Euler solver and a time resolution of $\tau = 1$ hour. The velocity dynamics network $f_v$ consists of three 2D ResNet blocks followed by a self-attention layer. The EEI module is implemented as a 3D ResNet, which takes concatenated heat fluxes as inputs to predict $\hat{Q}(t)$. The model is trained for 50 epochs using the AdamW optimizer with a learning rate of $5 \times 10^{-4}$, cosine decay scheduler, and batch size of 16. To balance computational cost and enable longer forecasting horizons, all variables are downsampled to $5.625^\circ$ spatial resolution and aligned to 6-hour intervals, then standardized before training and evaluation. All experiments are conducted on 4 NVIDIA Tesla V100 GPUs.

\subsection{A.4 Metrics}
We assess model performance using Mean Squared Error (MSE), Mean Absolute Error (MAE), and Anomaly Correlation Coefficient (ACC). All predictions are de-normalized before evaluation. Metrics are computed over the entire test set across both spatial and temporal dimensions. 
\begin{equation}
\text{MSE} = \frac{1}{N} \sum_{i=1}^{N} \left( \hat{y}_i - y_i \right)^2
\end{equation}

\begin{equation}
\text{MAE} = \frac{1}{N} \sum_{i=1}^{N} \left| \hat{y}_i - y_i \right|
\end{equation}

\begin{equation}
\text{ACC} = \frac{ \sum_{i=1}^{N} \left( \hat{y}i - \overline{\hat{y}} \right) \left( y_i - \overline{y} \right) } { \sqrt{ \sum{i=1}^{N} \left( \hat{y}i - \overline{\hat{y}} \right)^2 } \sqrt{ \sum{i=1}^{N} \left( y_i - \overline{y} \right)^2 } }
\end{equation}

Here, $\hat{y}_i$ and $y_i$ denote the predicted and ground truth SST values at the $i$-th spatiotemporal grid point, respectively; $N$ is the total number of evaluated points; and $\overline{\hat{y}}$, $\overline{y}$ represent the sample means of the predicted and ground truth values over the test period.

\subsection{A.5 Evaluation on Regional Forecasting}
Table~\ref{tab:regional-pacific}, Table~\ref{tab:regional-atlantic}, and Table~\ref{tab:regional-southern} report detailed regional forecasting metrics (MSE, MAE, ACC) over the Equatorial Pacific, Northwest Atlantic, and Southern Ocean, respectively. SSTODE consistently achieves the best performance across all regions and forecast horizons ($q$ = 5 and $q$ = 12) on the OceanVP benchmark. These results underscore the importance of integrating advection-diffusion dynamics and energy exchange modeling for robust regional forecasting.

\begin{table}[!h]
\centering
\small
\setlength{\tabcolsep}{3pt}
\renewcommand{\arraystretch}{0.8}
\begin{tabular}{l|ccc|ccc}
\toprule
\multirow{2}{*}{Model} & \multicolumn{3}{c|}{$q$ = 5} & \multicolumn{3}{c}{$q$ = 12} \\
 & MSE ↓ & MAE ↓ & ACC ↑ & MSE ↓ & MAE ↓ & ACC ↑ \\
\midrule
ConvLSTM     & 0.5578 & 0.0170 & 0.9721 & 1.1256 & 0.0245 & 0.9487 \\
MAU          & 0.5590 & 0.0174 & 0.9715 & 0.8755 & 0.0214 & 0.9579 \\
PredRNNv2    & 0.9216 & 0.0226 & 0.9541 & 1.2223 & 0.0264 & 0.9398 \\
TAU          & 0.5736 & 0.0178 & 0.9714 & 0.9200 & 0.0227 & 0.9556 \\
SimVPv2      & 0.5160 & 0.0166 & 0.9745 & 0.9612 & 0.0229 & 0.9533 \\
COTERE       & 0.5736 & 0.0178 & 0.9714 & 0.8653 & 0.0218 & 0.9578 \\
ClimODE      & 0.6064 & 0.0174 & 0.9705 & 0.8466 & 0.0210 & 0.9596 \\
\textbf{SSTODE} & \textbf{0.5157} & \textbf{0.0157} & \textbf{0.9748} & \textbf{0.7473} & \textbf{0.0193} & \textbf{0.9649} \\
\bottomrule
\end{tabular}
\caption{Regional forecasting results in the Equatorial Pacific (MSE $\times 10^3$ ↓, MAE ↓, ACC ↑) at $q$ = 5 and $q$ = 12.}
\label{tab:regional-pacific}
\end{table}

\begin{table}[!h]
\centering
\small
\setlength{\tabcolsep}{3pt}
\renewcommand{\arraystretch}{0.8}
\begin{tabular}{l|ccc|ccc}
\toprule
\multirow{2}{*}{Model} & \multicolumn{3}{c|}{$q$ = 5} & \multicolumn{3}{c}{$q$ = 12} \\
 & MSE ↓ & MAE ↓ & ACC ↑ & MSE ↓ & MAE ↓ & ACC ↑ \\
\midrule
ConvLSTM     & 1.4052 & 0.0205 & 0.9971 & 2.9191 & 0.0305 & 0.9940 \\
MAU          & 1.1125 & 0.0183 & 0.9977 & 1.8058 & 0.0245 & 0.9962 \\
PredRNNv2    & 1.9601 & 0.0251 & 0.9959 & 2.6103 & 0.0297 & 0.9945 \\
TAU          & 1.0621 & 0.0175 & 0.9978 & 1.8038 & 0.0235 & 0.9962 \\
SimVPv2      & 1.1709 & 0.0177 & 0.9977 & 1.9543 & 0.0258 & 0.9959 \\
COTERE       & 1.1118 & 0.0175 & 0.9977 & 1.7515 & 0.0231 & 0.9963 \\
ClimODE      & 1.1234 & 0.0167 & 0.9977 & 1.7046 & 0.0217 & 0.9963 \\
\textbf{SSTODE} & \textbf{0.9432} & \textbf{0.0152} & \textbf{0.9982} & \textbf{1.6445} & \textbf{0.0212} & \textbf{0.9964} \\
\bottomrule
\end{tabular}
\caption{Regional forecasting results in the Northwest Atlantic (MSE $\times 10^3$ ↓, MAE ↓, ACC ↑) at $q$ = 5 and $q$ = 12.}
\label{tab:regional-atlantic}
\end{table}

\begin{table}[!h]
\centering
\small
\setlength{\tabcolsep}{3pt}
\renewcommand{\arraystretch}{0.8}
\begin{tabular}{l|ccc|ccc}
\toprule
\multirow{2}{*}{Model} & \multicolumn{3}{c|}{$q$ = 5} & \multicolumn{3}{c}{$q$ = 12} \\
 & MSE ↓ & MAE ↓ & ACC ↑ & MSE ↓ & MAE ↓ & ACC ↑ \\
\midrule
ConvLSTM     & 0.4766 & 0.0107 & 0.9987 & 1.1772 & 0.0167 & 0.9970 \\
MAU          & 0.3504 & 0.0093 & 0.9990 & 0.7103 & 0.0135 & 0.9980 \\
PredRNNv2    & 0.6236 & 0.0126 & 0.9983 & 0.9231 & 0.0153 & 0.9975 \\
TAU          & 0.3014 & 0.0083 & 0.9991 & 0.6135 & 0.0119 & 0.9982 \\
SimVPv2      & 0.3051 & 0.0083 & 0.9991 & 0.6946 & 0.0138 & 0.9980 \\
COTERE       & 0.3259 & 0.0082 & 0.9991 & 0.6183 & 0.0118 & 0.9983 \\
ClimODE      & 0.3098 & 0.0069 & 0.9991 & 0.5716 & 0.0100 & 0.9984 \\
\textbf{SSTODE} & \textbf{0.2641} & \textbf{0.0063} & \textbf{0.9994} & \textbf{0.5597} & \textbf{0.0100} & \textbf{0.9984} \\
\bottomrule
\end{tabular}
\caption{Regional forecasting results in the Southern Ocean (MSE $\times 10^3$ ↓, MAE ↓, ACC ↑) at $q$ = 5 and $q$ = 12.}
\label{tab:regional-southern}
\end{table}

\subsection{A.6 Evaluation on Longer Horizon and Sparse Sampling}
Table~\ref{tab:temporal-generalization} reports detailed evaluation results under longer forecasting horizons (7 days) with two different temporal resolutions: 6-hour ($q=28$) and 12-hour ($q=14$) intervals on OceanVP. SSTODE also achieves remarkably stable results across all metrics and settings,  outperforming representative baselines. This robustness stems from the integration of energy exchange and diffusion-aware dynamics, which improve temporal consistency and suppress long-term error propagation--crucial for real-world ocean forecasting.

\begin{table}[!h]
\centering
\small
\setlength{\tabcolsep}{3pt}
\renewcommand{\arraystretch}{0.8}
\begin{tabular}{lccc|ccc}
\toprule
\multirow{2}{*}{Model} & \multicolumn{3}{c|}{6h ($q$ = 28)} & \multicolumn{3}{c}{12h ($q$ = 14)} \\
& MSE ↓ & MAE ↓ & ACC ↑ & MSE ↓ & MAE ↓ & ACC ↑ \\
\midrule
TAU        & 0.1847 & 0.2517 & 0.9968 & 0.1754 & 0.2344 & 0.9965 \\
SimVPv2    & 0.1800 & 0.2369 & 0.9966 & 0.1998 & 0.2789 & 0.9964 \\
COTERE     & 0.1619 & 0.2265 & 0.9969 & 0.1661 & 0.2334 & 0.9967 \\
ClimODE    & 0.1571 & 0.2144 & 0.9971 & 0.1645 & 0.2298 & 0.9967 \\
\textbf{SSTODE} & \textbf{0.1528} & \textbf{0.2084} & \textbf{0.9971} & \textbf{0.1579} & \textbf{0.2170} & \textbf{0.9969} \\
\bottomrule
\end{tabular}
\caption{Temporal generalization results on OceanVP under 6-hour ($q$ = 28) and 12-hour ($q$ = 14) settings over a 7-day horizon.}
\label{tab:temporal-generalization}
\end{table}

\subsection{A.7 Evaluation on Higher Spatial Resolution}
Table~\ref{tab:era5-resolution} reports SST forecasting performance at the $2.8125^\circ$ spatial resolution on the ERA5 benchmark. SSTODE achieves the lowest MSE and MAE, confirming its ability to scale effectively to higher spatial resolutions. Due to computational constraints, experiments are conducted at moderate resolution. Nevertheless, the results demonstrate the scalability of SSTODE and its potential for extension to finer spatial grids (e.g., $0.25^\circ$), enabling more accurate modeling of mesoscale and submesoscale ocean dynamics.

\begin{table}[!h]
\centering
\setlength{\tabcolsep}{6pt}
\renewcommand{\arraystretch}{0.8}
\begin{tabular}{lccc}
\toprule
Model & MSE ↓ & MAE ↓ & ACC ↑ \\
\midrule
TAU        & 0.0250 & 0.0826 & 1.0000 \\
SimVPv2    & 0.0230 & 0.0810 & 1.0000 \\
COTERE     & 0.0278 & 0.0947 & 1.0000 \\
ClimODE    & 0.0210 & 0.0706 & 1.0000 \\
\textbf{SSTODE} & \textbf{0.0190} & \textbf{0.0580} & \textbf{1.0000} \\
\bottomrule
\end{tabular}
\caption{Forecasting results on ERA5 ($64 \times 128$ grid, 6-hour interval, $q$ = 5). SSTODE achieves the best accuracy across all metrics.}
\label{tab:era5-resolution}
\end{table}

\subsection{A.8 Evaluation on Model Parameters Comparison}

Table~\ref{tab:model-params} compares model sizes across baselines under the same $2.8125^\circ$-resolution SST forecasting setting. 
Compared to both recurrent-based and recurrent-free baselines, our model demonstrates superior parameter efficiency. While SSTODE and ClimODE adopt the autoregressive formulation with comparable lightweight model sizes (4.62M vs 3.67M), SSTODE achieves a significant $\sim$10\% improvement in forecasting accuracy (0.019 vs 0.021 in MSE), demonstrating a favorable trade-off between performance and model complexity.

\begin{table}[!h]
\centering
\setlength{\tabcolsep}{10pt}
\renewcommand{\arraystretch}{0.8}
\begin{tabular}{lc}
\toprule
\textbf{Models} & \textbf{Params.} \\
\midrule
MAU            & 11.75M \\
PredRNNv2      & 23.59M \\
TAU            & 9.39M \\
SimVPv2          & 9.80M \\
ClimODE        & 3.67M \\
\textbf{SSTODE (Ours)} & \textbf{4.62M} \\
\bottomrule
\end{tabular}
\caption{Comparison of model sizes in parameter counts.}
\label{tab:model-params}
\end{table}

\subsection{A.9 Comparison with Neural Operator Baselines.}

\begin{table}[h!]
\centering
\begin{tabular}{lccc}
\toprule
\textbf{Models} & \textbf{MSE ↓} & \textbf{MAE ↓} & \textbf{ACC ↑} \\
\midrule
\multicolumn{4}{l}{\textit{FNO-based}} \\
FNO           & 0.0750 & 0.1564 & 0.9987 \\
U-FNO         & 0.0745 & 0.1512 & 0.9987 \\
U-NO          & 0.1331 & 0.2361 & 0.9981 \\
F-FNO         & 0.0725 & 0.1469 & 0.9987 \\
LSM           & 0.0809 & 0.1643 & 0.9986 \\
\midrule
\multicolumn{4}{l}{\textit{Transformer-based}} \\
Galerkin      & 0.0771 & 0.1528 & 0.9986 \\
Factformer    & 0.0857 & 0.1938 & 0.9986 \\
Transolver    & 0.0898 & 0.1953 & 0.9984 \\
\midrule
\textbf{SSTODE} & \textbf{0.0527} & \textbf{0.1107} & \textbf{0.9990} \\
\bottomrule
\end{tabular}
\caption{Comparison with neural operator baselines on OceanVP (5-step horizon). The best results are highlighted in bold.}
\label{tab:neural-operator}
\end{table}

We further benchmarked our model against state-of-the-art neural operator approaches, including both Fourier-based and transformer-based architectures implemented in the \texttt{Neural-Solver Library}\footnote{\url{https://github.com/thuml/Neural-Solver-Library}}. 
Specifically, we evaluated Fourier Neural Operator (FNO)~\citep{lifourier}, its variants U-FNO \cite{wen2022u} and F-FNO~\citep{tranfactorized}, the U-shaped Neural Operator (U-NO)~\citep{rahmanu}, and the Latent Spectral Model (LSM)~\citep{wu2023solving}. 
For transformer-based operator learning, we included Galerkin Transformer~\citep{cao2021choose}, Factformer~\citep{li2023scalable}, and Transolver~\citep{wu2024transolver}. 
All models were trained and evaluated on the OceanVP dataset with a 5-step forecasting horizon using consistent data preprocessing and normalization.

As shown in Table~\ref{tab:neural-operator}, our model achieves the best performance across all metrics (MSE, MAE, and ACC), outperforming both FNO- and transformer-based neural operator baselines. 
This demonstrates the advantage of introducing explicit advection–diffusion constraints and continuous-time dynamics within a Neural ODE framework, enabling better physical consistency and forecasting stability.

\end{document}